\documentclass[sigconf]{acmart}
\usepackage{algorithmicx,algorithm}
\usepackage{subfigure}
\usepackage{multirow}
\usepackage{pifont}
\usepackage{paralist}
\usepackage{balance}

\usepackage{titlesec}
\titleformat{\section}{\normalfont\Large\bfseries\MakeUppercase}{\thesection\ }{0.5em}{}
\titlespacing*{\subsection}{0pt}{0.3\baselineskip}{0.4\baselineskip}

\AtBeginDocument{%
  \providecommand\BibTeX{{%
    \normalfont B\kern-0.5em{\scshape i\kern-0.25em b}\kern-0.8em\TeX}}}

\copyrightyear{2023}
\acmYear{2023}
\setcopyright{acmlicensed}\acmConference[MM '23]{Proceedings of the 31st ACM International Conference on Multimedia}{October 29-November 3, 2023}{Ottawa, ON, Canada}
\acmBooktitle{Proceedings of the 31st ACM International Conference on Multimedia (MM '23), October 29-November 3, 2023, Ottawa, ON, Canada}
\acmPrice{15.00}
\acmDOI{10.1145/3581783.3612065}
\acmISBN{979-8-4007-0108-5/23/10}

\acmSubmissionID{1669}

\title{Versatile Face Animator: Driving Arbitrary 3D Facial Avatar \\in RGBD Space}

\author{Haoyu Wang}
\email{why22@mails.tsinghua.edu.cn}
\affiliation{%
  \institution{Department of Computer Science and \\
Technology, Tsinghua University}
  \city{Beijing 100084}
  \country{China}
  \postcode{100084}}

\author{Haozhe Wu}
\email{wuhz19@mails.tsinghua.edu.cn}
\affiliation{%
  \institution{Department of Computer Science and \\
Technology, Tsinghua University}
  \city{Beijing 100084}
  \country{China}
  \postcode{100084}}

\author{Junliang Xing}
\email{jlxing@tsinghua.edu.cn}
\affiliation{%
  \institution{Department of Computer Science and \\
Technology, Tsinghua University}
  \city{Beijing 100084}
  \country{China}
  \postcode{100084}}

\author{Jia Jia}
\email{jjia@tsinghua.edu.cn}
\affiliation{%
  \institution{Department of Computer Science and \\
Technology, Tsinghua University\\Beijing National Research Center for \\ Information Science and Technology}
  \city{Beijing 100084}
  \country{China}
  \postcode{100084}}
\authornote{Corresponding author}

\def\authors{Haoyu Wang, Haozhe Wu, Junliang Xing, Jia Jia}

\renewcommand{\shortauthors}{Haoyu Wang, Haozhe Wu, Junliang Xing, \& Jia Jia}

\renewcommand{\shorttitle}{Versatile Face Animator: Driving Arbitrary 3D Facial Avatar in RGBD Space}

\begin{document}

\begin{abstract}

Creating realistic 3D facial animation is crucial for various applications in the movie production and gaming industry, especially with the burgeoning demand in the metaverse. However, prevalent methods such as blendshape-based approaches and facial rigging techniques are time-consuming, labor-intensive, and lack standardized configurations, making facial animation production challenging and costly. In this paper, we propose a novel self-supervised framework, Versatile Face Animator, which combines facial motion capture with motion retargeting in an end-to-end manner, eliminating the need for blendshapes or rigs. 
Our method has the following two main characteristics: 1) we propose an RGBD animation module to learn facial motion from raw RGBD videos by hierarchical motion dictionaries and animate RGBD images rendered from 3D facial mesh coarse-to-fine, enabling facial animation on arbitrary 3D characters regardless of their topology, textures, blendshapes, and rigs; and 2) we introduce a mesh retarget module to utilize RGBD animation to create 3D facial animation by manipulating facial mesh with controller transformations, which are estimated from dense optical flow fields and blended together with geodesic-distance-based weights.
Comprehensive experiments demonstrate the effectiveness of our proposed framework in generating impressive 3D facial animation results, highlighting its potential as a promising solution for the cost-effective and efficient production of facial animation in the metaverse.

\end{abstract}

\ccsdesc[500]{Computing methodologies~Animation}

\keywords{ 3D facial animation; Motion capture; Motion retargeting.}

\maketitle

\section{Introduction}
\label{intro}
Creating 3D facial animation is a complex task with numerous applications in fields such as movie production and the gaming industry \cite{chandran2022local}. With the increasing popularity of the \textit{metaverse}, the demand for 3D facial animation has significantly grown. In the metaverse, users expect to interact with diverse avatars, including themselves, celebrities, and fictional characters like the \textit{Na'vi} in \textit{Avatar}. These needs necessitate sophisticated facial animation techniques that accurately capture the nuances of human expressions and emotions. However, creating believable facial animation remains a challenge even for the most skilled animators, Hollywood filmmakers, or game developers due to the high level of expertise and the significant amount of time required.

One of the standard approaches for creating 3D facial animation in the industry is \textit{performance retargeting}. This method involves capturing facial motion from a real actor and transferring that motion to a target 3D avatar \cite{chandran2022local}. The most common technique for achieving this is the blendshape-based methods \cite{lewis2014practice,carrigan2020expression,li2010example}. Blendshapes consist of predefined deformations of the facial mesh, which can be combined to create a wide range of facial expressions using various input weights. This process allows for transferring facial motion from a source video to a target character, even if visually dissimilar. However, creating blendshapes with high flexibility and rich expressiveness is time-consuming and labor-intensive, as it may require hundreds of various expressions for a single character. For example, in \textit{The Curious Case of Benjamin Button}, filmmakers captured 170 blendshapes of Brad Pitt \cite{flueckiger2011computer}. Additionally, the absence of a standard configuration for creating blendshapes complicates cross-mapping between different expression spaces and hinders motion transfer across distinct avatars.
Another prevalent approach for generating 3D facial animation is facial rigging, which involves manipulating motion controls to create the desired animation \cite{orvalho2012facial}. However, facial rigging is typically an iterative and laborious process since no consistent rig can be used for all possible motions. As a result, the rigging process often becomes a bottleneck in 3D animation production. Furthermore, varying standards across different software packages make transferring facial motion across characters with distinct rigs extremely challenging.

Both blendshape-based methods and facial rigging, as discussed above, are facing similar challenges: (i) they are time-consuming and labor-intensive, which limits their accessibility to common users, and (ii) they lack standardization, making it challenging to transfer facial motion across different characters with varying rigs or blendshape configurations. These challenges hinder the development of the metaverse, where users expect to act on arbitrary characters in a short set-up time. Motivated by these issues, we aim to explore a new solution that directly drives the facial mesh with raw RGBD videos, eliminating reliance on blendshapes or rigs. 

To this end, we propose a novel framework, Versatile Face Animator (VFA), that combines facial motion capture with motion retargeting to drive the facial mesh with captured RGBD videos end-to-end. We aim to model the facial motion in color and depth fields and generate RGBD animation to drive facial mesh. To achieve this goal, our framework consists of the RGBD animation module and the mesh retarget module. First, the RGBD animation module generates the animated RGBD frame with hierarchical motion dictionaries. It then estimates the correspondence between the source RGBD image and the animated frame with a distilled flow generator. More specifically, the RGBD animation module encodes arbitrary facial motion into a combination of basic transformations in the motion dictionary and generates the animated frame from coarse to fine. The flow generator is then trained to estimate a dense optical flow field for building correspondence between source images and animated frames. The flow generator is distilled from the RGBD generator under animated RGBD frames' supervision, eliminating the need for extra labels. The mesh retarget module then deforms the facial mesh with the dense optical flow. It first detects the controlling points of the mesh automatically and then calculates the geodesic-distance-based controlling weights of each vertex. Afterward, the mesh retarget module estimates controlling point transformations according to the dense optical flow. The transformations are then blended with calculated weights to deform the mesh.

To summarize, this work makes three main contributions:
\begin{compactitem}
    \item We propose VFA, a novel self-supervised framework that combines facial motion capture with facial motion retargeting in an end-to-end manner, providing a cost-effective solution for 3D facial animation production.
    \item We introduce a new method to learn facial motion in both the color field and the depth field with hierarchical motion dictionaries and generate RGBD animation coarse-to-fine.
    \item We present a new pipeline for transferring RGBD animation to create 3D animation by deforming the mesh with controller transformations, which are estimated from a dense optical flow field and blended with geodesic-distance-based controlling weights.
\end{compactitem}

Our approach presents two main advantages: 1) It employs self-supervised training using raw facial RGBD data, eliminating the need for annotation or additional configuration; and 2) it can animate arbitrary 3D characters, regardless of their topology, blendshapes, or rigs. A comprehensive set of experiments, encompassing both qualitative and quantitative analyses, showcases the outstanding performance of our method in generating 3D facial animations at a relatively low cost. This positions our approach as a promising solution for 3D facial animation production.

 \section{Related Work}
\textbf{Blendshapes and Facial Rigging.} Blendshapes, an approximate semantic parameterization of facial expression, have become the standard approach to generating realistic facial animation in the industry \cite{lewis2014practice}. With little computation, an extensive range of expressions can be expressed by linearly combining blendshape targets. However, hundreds of blendshape targets are required to build an expression space with enough expressiveness. To reduce this unbearable cost, researchers have proposed methods to reduce the demands of training expressions \cite{carrigan2020expression} or to fine-tune the blendshape model based on a generic prior \cite{li2010example}. To deal with transferring blendshape weights across different expression spaces, Kim \textit{et al.} proposed a method that animated rendered images in the 2D domain and then estimated blendshape weights from the retargeted images \cite{kim2021deep}, which is similar to our proposed framework but can only drive a particular set of avatars due to the reliance of blendshapes.
Facial rigging is another widely used technique that seeks to build motion controls and animate the target character \cite{orvalho2012facial}. To some extent, blendshape weights can be seen as a kind of control rig. Several neural approaches have been proposed to estimate facial rigs from animation using neural networks \cite{bailey2020fast, song2020accurate, zhao2022uncertainty}, which enables motion transfer across characters. However, both blendshape-based methods and facial rigging techniques still suffer from their high configuration costs and lack of standard criteria in production.
In our framework, we aim to model the facial motion of RGBD frames via 2D facial animation methods. This approach eliminates the need for blendshapes or facial rigs, reducing the configuration cost while maintaining satisfactory performance.
\begin{figure*}[t]
\centering
    \includegraphics[width=\textwidth]{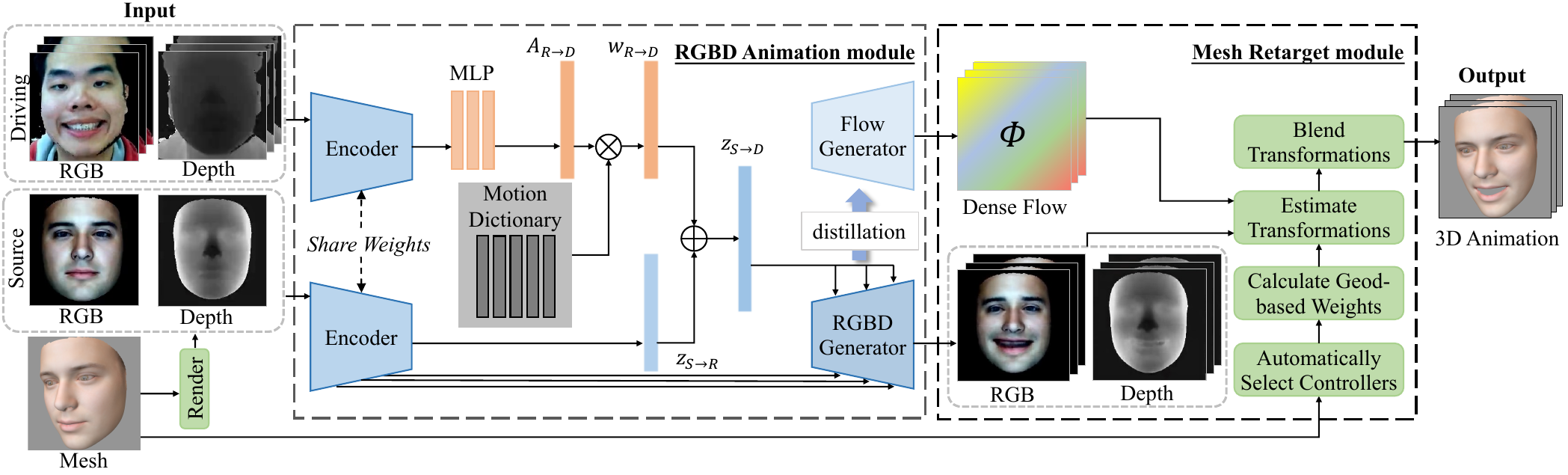}
    \caption{An overview of our proposed framework VFA. We generate 3D facial animation with the source mesh and captured RGBD video as input in an end-to-end manner. Our model consists of an RGBD animation module and a mesh retarget module. The RGBD animation module encodes source image S to $z_{S\rightarrow R}$ and encodes facial motion from driving frame D to $w_{R\rightarrow D}$ using the motion dictionary $\mathcal{D}_m$. With the composed latent code $z_{S\rightarrow D}$, the RGBD animation module generates the driven RGBD frame and estimates a dense optical flow field $\Phi$, which can be used to warp the source image. The mesh retarget module then warp the source mesh $\mathcal{S}$ utilizing information from the animated RGBD pair and dense flow $\Phi$ to generate 3D facial animation.}
    \label{fig:overview}
    \vspace{-4mm}
\end{figure*}

\textbf{Data-driven 3D Facial Retargeting.}  Facial retargeting techniques have significantly advanced with the development of neural networks. The Variational Auto Encoder (VAE) \cite{kingma2013auto} has been introduced to disentangle facial identity and expression in the latent space, allowing for the transfer of facial motion \cite{tran2018nonlinear, ranjan2018generating}.  Recently, Zhang \textit{et al.} \cite{zhang2020facial} proposed training character-specific VAE models to transfer characters' expressions across different domains.  Most current studies on neural facial retargeting methods are based on the 3D morphable model (3DMM), which isolates identity and expressions \cite{ li2017learning, feng2021learning}. Moser \textit{et al.} \cite{moser2021semi} inspired our work by proposing to treat 3D facial retargeting as 2D face swapping between the actor and the target character. They animated the rendered images using an unsupervised image-to-image translation model and then regressed the 3DMM parameters from the animated images. However, these methods typically require a large amount of paired high-accuracy 3D facial data, which is difficult to capture. Additionally, 3DMM-based methods suffer from a lack of expressiveness due to their linear nature.
In our work, we propose to train our method with RGBD videos which be captured easily using a single Azure Kinect V2 camera. Our method eliminates the need for paired 3D facial data and allows arbitrary deformation by warping the facial mesh with an estimated optical flow field.

\textbf{2D Facial Animation.} Generating 2D facial animation, also known as face reenactment, has seen rapid progress due to advancements in deep learning. To facilitate the image-to-image translation model for facial reenactment, researchers have introduced facial structure representations as prior knowledge, such as facial landmarks,  \cite{wang2018video, yang2020transmomo, zakharov2019few, chen2022d2animator}, semantic label maps \cite{pan2019video, wang2019few} and optical flows \cite{ohnishi2018hierarchical, li2018flow, yi2020animating}. However, such semantic labels for supervised learning are usually difficult to access for training and inference. A self-supervised motion transfer approach, i.e., the first-order motion model, was introduced to learn facial keypoint transformations from raw videos and warp the source image with the estimated dense optical flow \cite{siarohin2019first}. Based on the first-order motion model, Hong \textit{et al.} proposed to recover facial depth images in a self-supervised manner and leverage the depth information to generate 2D facial animation \cite{hong2022depth}. Wang \textit{et al.} proposed a  novel method called LIA to drive still 2D images via latent space navigation, which eliminates the need for explicit structure representations like keypoints and can discover high-level motion transformations in latent space \cite{wang2022latent}.  However, there remains a gap between 2D face reenactment and 3D facial retargeting, since most methods treat 3D information as prior knowledge, and few focus on how to transfer facial motion in the 3D mesh. We bridge the gap by incorporating the depth information from RGBD videos and modeling facial motion in the depth field using the depth motion dictionary to generate animated RGBD frames and subsequently deform the facial mesh.
\vspace{-4mm}
\section{Methodology}

We aim to animate the source 3D facial mesh $\mathcal{S}$ of a target avatar based on the facial motion from a raw RGBD video $\mathcal{D}$ captured by Azure Kinect V2. To achieve this, our proposed end-to-end framework consists of the RGBD animation module and the mesh retarget module, as depicted in Fig. \ref{fig:overview}. 

\textbf{The RGBD animation module} is designed to model facial motion extracted from the driving frame $\textbf{D}$ of video $\mathcal{D}$ and transfer it to the rendered image $\textbf{S}$ from mesh $\mathcal{S}$.  Additionally, the RGBD animation module estimates a dense optical flow field $\Phi$, which can be used to establish correspondence between the source image and the animated frame. The estimated dense flow $\Phi$ will be utilized in the mesh retarget module.

\textbf{The mesh retarget module} deforms the source mesh $\mathcal{S}$ with detected facial landmarks as controllers and geodesic-distance-based controlling weights. The transformations of controllers are estimated using dense flow field $\Phi$ from the RGBD animation module and then are mapped to 3D world space with the animated depth frames. Finally, the transformations are blended to generate the desired 3D facial animation frame by frame. In the following, we will introduce the two modules in detail.

\subsection{RGBD Animation Module}

\begin{figure*}[t]
\centering
    \includegraphics[width=1.0\linewidth]{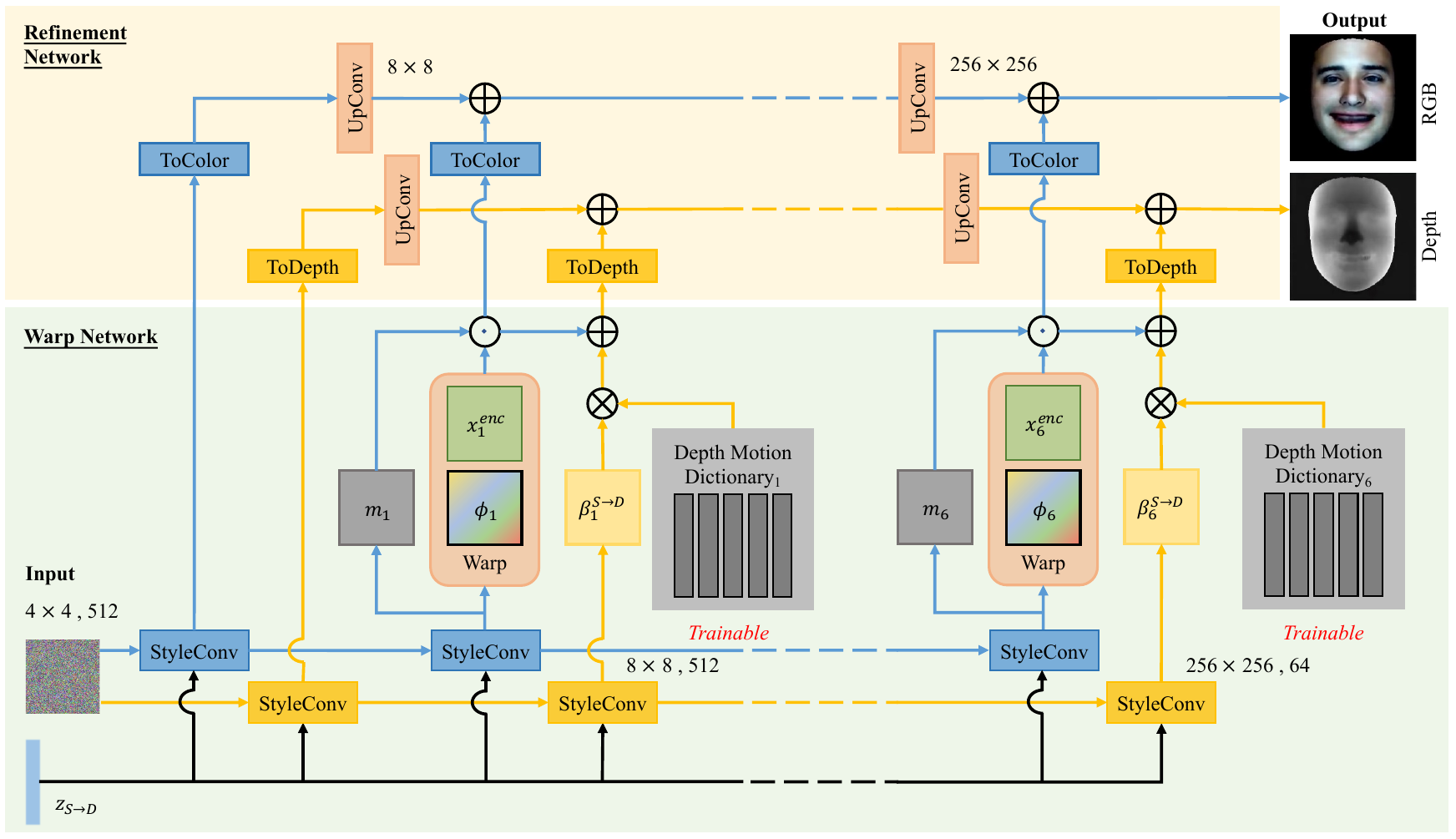}
    \caption{An overview of the generator. The generator employs a 6-level pyramid architecture, comprising the warp network and the refinement network. The warp network utilizes \textit{StyleConv} \cite{karras2020analyzing} to estimate optical flow fields $\{\phi_i\}_1^6$ and masks $\{m_i\}_1^6$, and warps feature maps $x_i^{enc}$ from the encoder. The depth motion dictionaries are introduced to model motion in the depth field. The refinement network then utilizes convolution layers to generate the animated frames in a coarse-to-fine manner. }
    \label{fig:generator}
    \vspace{-3mm}
\end{figure*}

\subsubsection{Encoder}
The RGBD animation module is inspired by the LIA method proposed by Wang \textit{et al.} \cite{wang2022latent}. It utilizes an auto-encoder structure and consists of an encoder and two generators. The encoder encodes images to latent codes, and the RGBD generator decodes these codes and generates animated RGBD frames coarse-to-fine. Furthermore, the flow generator estimates dense optical flow fields which will be utilized in the mesh retarget module. In the following section, we proceed to discuss them comprehensively.

The encoder is designed to learn a latent code $z_{S\to D}$ to represent the motion transformation from \textbf{S} to \textbf{D}. However, as Wang \textit{et al.} \cite{wang2022latent} point out, it is challenging to learn  $z_{S\to D}$ directly from the input image pair, as the model needs to model the direction and norm of the vector $z_{S\to D}$ simultaneously. To overcome this challenge, we assume there exists a reference frame \textbf{R} so that the motion transformation from \textbf{S} to \textbf{D} can be decomposed as $\textbf{S}\to\textbf{R}\to\textbf{D}$. This allows us to learn the transformations $\textbf{S}\to\textbf{R}$ and $\textbf{R}\to\textbf{D}$ independently and then compose them to represent $\textbf{S}\to\textbf{D}$. We model $z_{S\to D}$ as the target point in the latent space, which can be reached from the source point $z_{S\to R}$ along a path $w_{R\to D}$ in the latent space. Mathematically, the latent code $z_{S\to D}$ can be decomposed as $z_{S\to D}=z_{S\to R}+w_{R\to D}
$.


To ensure that the learned latent codes are in the same latent space, we utilize a single encoder to encode the source image and the driving image. As depicted in Fig. \ref{fig:overview}, the encoder encodes the source image and the driving image as $z_{S\to R}$ and $z_{D\to R}$ respectively. To extract high-level motion information from $z_{D\to R}$, we propose to encode motion via Linear Motion Decomposition \cite{wang2022latent}. Specifically, we introduce a learnable orthogonal basis called the motion dictionary ${D_m}$. Each vector of the motion dictionary represents a direction $\mathbf{d_i}$ of the motion space. $z_{D\to R}$ is mapped to a magnitude vector $A_{R\to D}$ by an MLP layer. Then, the latent path $w_{R\to D}$ is obtained by linearly combining the magnitude vector $A_{R\to D}$ with the basis vector $\mathbf{d_i}$ from the motion dictionary $D_m$. With the latent code $z_{S\to R}$ learned from the source image \textbf{S} and the latent path $w_{R\to D}$ extracted from the driving frame \textbf{D}, we can obtain $z_{S\to D}$ which represents the transformation $\textbf{S}\to\textbf{D}$.

\subsubsection{Generator}
\label{subsubsection:gen}

We proceed to introduce the RGBD generator and the flow generator respectively.

The general architecture of the RGBD generator is depicted in Fig. \ref{fig:generator}, which consists of the flow and refinement networks. To learn multi-scale features, the generator employs a 6-level pyramid architecture and uses skip connection between layers.  The \textit{StyleConv} \cite{karras2020analyzing} layer is introduced to decode $z_{S\to D}$ and estimate multiple levels of optical flow fields $\{\phi_i\}_1^6$. These optical flow fields $\{\phi_i\}_1^6$ are then used to warp the feature maps $x_i^{enc}$ from the corresponding level of the source encoder. However, as Siarohin \textit{et al.} \cite{siarohin2019first} pointed out, the occluded parts of the source image \textbf{S} can not be recovered by simply warping the image. Consequently, we propose to estimate the masks $\{m_i\}_1^6$ along with $\{\phi_i\}_1^6$. The masks are utilized to mask the occluded region to inpaint in the refinement network. In this way, the transformed feature map is formulated as:
\begin{equation*}
x_i'=m_i \odot f_w(x_i^{enc}, \phi_i),
\end{equation*}
where $f_{w}$ represents the backward warping function. 

The estimated optical flow $\{\phi_i\}_1^6$ provides the pixel-wise correspondence between the source and warped images in the 2D domain, but it is not enough to model motion in the depth field. When an object moves relative to the camera's z-axis, it can alter the pixel values in the depth frame, which can not be captured by the optical flow $\{\phi_i\}_1^6$.  To address this issue, we introduce the depth motion dictionaries $\{D^{depth}_{i}\}_1^6$ to adequately learn motion in the depth field. The basis vectors of the depth motion dictionary represent the direction of the depth motion space. By linearly combining the basis vectors with the predicted magnitude vector $\beta_i^{S\to D}$, we estimate the motion in the depth field. This allows us to obtain the feature map $x_i^{depth}$ for generating accurate depth images. $x_i^{depth}$ can be expressed as:
\begin{equation*}
    \vspace{-5mm}
    x_i^{depth}=m_i \odot f_w(x_i^{enc}, \phi_i) + \sum_{j=1}^{M} \beta_{i,j}^{S\to D}\textbf{d}_{i,j}^{depth},
\end{equation*}
where $M$ denotes the size of the depth motion dictionary $D^{depth}_{i}$,  and $\textbf{d}_{i,j}^{depth}$ represents the basic vectors of $D^{depth}_{i}$.

In the refinement network, we adopt a coarse-to-fine approach to generate precise RGBD results. At each layer of the refinement network, we combine the upsampled results from the previous layer with inpainted feature maps to generate images.
This iterative process allows us to progressively refine the generated images in a hierarchical manner, capturing finer details and improving the overall visual quality of the outputs.

We note that the warp network predicts optical flow fields $\phi_i$ to warp the feature maps $x_i^{enc}$ from the encoder. This poses a challenge for the mesh retarget module in analyzing the flow fields and accurately tracking the movement of controlling points during animation. To address this challenge, we introduce a dense flow generator to generate a dense optical flow field, denoted as $\Phi$, which represents the pixel-wise correspondence between the input image \textbf{S} and the animated image. The dense flow generator is trained through distillation from the original generator, utilizing the warped image from the refinement network and the source image as training data. This training scheme allows the dense flow generator to generate $\Phi$ without conversion or extra training data. This approach facilitates the mesh retarget module to track the transformation of controllers to perform mesh retargeting.

\subsection{Mesh Retarget Module}
\label{subsection:retarget}

The design of the mesh retarget module is inspired by linear blend skinning (LBS), the most popular shape deformation algorithm for real-time animation due to its efficiency and simplicity \cite{kavan2008lbs, kavan2014direct}. Our method modifies the vertex positions while preserving mesh connectivity to achieve accurate and consistent animation results for different target meshes. Furthermore, we determine controlling weights using geodesic distances, which maintain the mesh topology and produce natural results.

We use an open-source library, Mediapipe \cite{lugaresi2019mediapipe} to detect facial landmarks as controllers. These detected landmarks provide rich semantic information facilitating reasonable and transferable blend transformations. Then we compute geodesic distances on the mesh surface between controlling points and mesh vertices.  For each mesh vertex, we assign the 10 nearest controllers to determine the controlling weights based on the inverse square of the geodesic distances.
We must note that we use geodesic distance instead of Euclidean distance to preserve the mesh topology. Specifically, using geodesic distance as the metric ensures that the upper and lower lip vertices are not mistakenly considered neighbors. Further details on the comparison are discussed in Section \ref{subsection:ablation}.

When generating animation frame by frame, we analyze the flow generator's dense flow field $\Phi$ to estimate controller transformations. However, these estimated transformations are in the 2D screen space, while the source mesh $\mathcal{S}$ exists in 3D world space. The transformations can not be directly aggregated to deform the mesh. Therefore, to map the transformation to 3D space, we estimate the position of controller $v_j$ utilizing the depth of its corresponding pixel and unproject it to the 3D world space using the perspective matrix. We formulate this process as follows: 
\begin{equation*}
    \textbf{v}_j=\textbf{P}^{-1}(v_j.x, v_j.y, d(v_j), 1)^T,
\end{equation*} 
where $P$ denotes the perspective matrix, and $d(v_j)$ denotes the depth value of $v_j$'s corresponding pixel in the image. We then track the movement of the controllers with the dense optical flow field $\Phi$ and estimate controller transformations in 3D world space. The deformed vertex positions can be calculated by linearly combining the transformations with the controlling weights.

The mesh retarget module plays a critical role in our proposed framework, bridging the 2D image animation problem and the 3D facial retargeting problem. By utilizing geodesic-determine controlling weights and incorporating depth information from generated frames, this module enables direct warping of the source mesh $\mathcal{S}$ without blendshapes or rigs. This integration simplifies retargeting facial motion to avatars, making our proposed framework a cost-efficient solution for creating realistic 3D facial animation.

\subsection{Training Losses}
\label{subsection:loss}

In the training stage, the RGBD animation module is trained in a self-supervised manner to reconstruct the driving frame \textbf{D}. To further enhance the robustness and performance of our model, we fine-tune the RGBD module based on the weights of LIA pre-trained in the VoxCeleb \cite{nagrani2017voxceleb} dataset. Four losses are used to train the RGBD module: a reconstruction loss $\mathcal{L}_{rec}$, a perceptual loss $\mathcal{L}_{vgg}$, a smooth loss $\mathcal{L}_{sm}$ and a structure preserve loss $\mathcal{L}_{sp}$.



$\mathcal{L}_{rec}$ is calculated using $\mathcal{L}_1$ distance, while the perceptual loss $\mathcal{L}_{vgg}$, proposed by Johnson \textit{et al.} \cite{johnson2016perceptual}, is calculated on multi-scales feature maps extracted from the pre-trained VGG-19 network \cite{simonyan2014very}. 



To improve the quality of the depth image we generated, we design two depth-related losses: the smooth loss $\mathcal{L}_{sm}$ and the structure preserve loss $\mathcal{L}_{sp}$. The smooth loss is designed based on the Laplacian operator to improve the smoothness of $\hat{\mathbf{D}}$:
\begin{equation*}
\mathcal{L}_{sm}=\mathbb{E}\big\vert  
\nabla^2\mathbf{D}-\nabla^2\hat{\mathbf{D}}  \big\vert_2,
\end{equation*}
where $\nabla^2\mathbf{D} $ denotes a Laplacian operator:$\nabla^2\mathbf{D}(x,y)=\mathbf{D}(x+1,y)+\mathbf{D}(x-1,y)+\mathbf{D}(x,y+1)+\mathbf{D}(x,y-1)-4\mathbf{D}(x,y)$

\begin{table*}[t]
\centering
\caption{\textbf{Results of Same-identity Reconstruction.} We compared our method with three state-of-the-art methods on two datasets: MMFace4D \cite{wu2023mmface4d} and VocCeleb \cite{nagrani2017voxceleb}. For all metrics except CSIM, the lower, the better.}
\setlength{\tabcolsep}{7pt}
    \begin{tabular}{l|ccccccc|ccccc}
    \toprule 
    \multirow{2}{*}{Method} & \multicolumn{7}{c|}{MMFace4D}                              & \multicolumn{5}{c}{VoxCeleb}         \\
        & $\mathcal{L}_1$    & LPIPS & AKD   & AED   & CSIM  & Depth $\mathcal{L}_1$ & $\mathcal{L}_{sm}$ & $\mathcal{L}_1$    & LPIPS & AKD   & AED   & CSIM  \\
        \cmidrule(lr){1-13}
    FOMM \cite{siarohin2019first}                  & 10.61    & 0.123 & 2.822 & 0.537 & 0.839 & 0.040    & 0.019  & 12.27     & 0.128 & 2.398 & 0.574 & 0.814 \\
    OSFV \cite{wang2021one}                  & 9.32     & 0.121 & 2.612 & 0.528 & 0.842 & 0.037    & 0.017  & 11.87     & 0.121 & \textbf{2.385} & 0.562 & 0.822 \\
    DaGAN \cite{hong2022depth}                   & 8.13     & 0.104 & 2.502 & 0.529 & \textbf{0.844} & 0.036    & 0.015  & 11.77     & 0.122 & 2.542 & 0.570 & 0.820 \\
    Ours                     & \textbf{8.04}     & \textbf{0.104} & \textbf{2.312} & \textbf{0.440} & \textbf{0.844} & \textbf{0.030}    & \textbf{0.015}  & \textbf{11.26}     & \textbf{0.119} & 2.475 & \textbf{0.570} & \textbf{0.820} \\
    \bottomrule
    \end{tabular}
    \label{tab:recon}
\end{table*}

Furthermore, to preserve the original geometric structure and the depth discontinuity along the edges in depth frames, we introduce the structure preserve loss $\mathcal{L}_{sp}$ proposed by Jeon \textit{et al.} \cite{jeon2018reconstruction}:
\begin{equation*}
    \mathcal{L}_{sp}=\mathbb{E}_p\Big\vert \max_{q\in\Omega(p)}\big\vert \nabla \mathbf{D}(p)\big\vert-\max_{q\in\Omega(p)}\big\vert \nabla \hat{\mathbf{D}}(p)\big\vert \Big\vert_2,
\end{equation*}
where $\Omega(p)$ denotes a local region in the neighborhood of $p$, and $\nabla \mathbf{D}(p)$ denotes the gradient calculated as: $\nabla_x \mathbf{D}(x,y)=\mathbf{D}(x+1,y)-\mathbf{D}(x-1,y),\nabla_y \mathbf{D}(x,y)=\mathbf{D}(x,y+1)-\mathbf{D}(x,y-1)$. In practice, we set $\Omega(p)$ as a $5\times5$ window near the pixel $p$.

Our full loss function while training the RGBD animation module is the combination of the four losses discussed above:
\begin{equation*}
    \mathcal{L}=\mathcal{L}_{vgg}+\lambda_{rec}\mathcal{L}_{rec}+\lambda_{sm}\mathcal{L}_{sm}+\lambda_{sp}\mathcal{L}_{sp},
\end{equation*}
where we use three user-define hyperparameters for balance. In practice, these parameters are set as $\lambda_{rec}=\lambda_{sm}=200,\lambda_{sp}=50$. It is important to note that our method exhibits robustness to different hyperparameter settings. Consequently, we do not demonstrate an ablation study examining the combined loss function.

\section{Experiments}


\subsection{Experiments Settings}

\textbf{Dataset} Our model is pre-trained in the VoxCeleb \cite{nagrani2017voxceleb} dataset, and fine-tuned in the \textbf{MMFace4D} dataset proposed by Wu \textit{et al.} \cite{wu2023mmface4d}. The MMFace4D dataset is a large-scale facial RGBD video dataset captured by Azure Kinect V2.  During training, we selected 191 identities, used 16,549 videos, cropped the facial region, removed the background, resized the frames to $256\times256$, and normalized the frames to the range of $[-1, 1]$.  For testing, we utilized the test dataset from VoxCeleb \cite{nagrani2017voxceleb} and VoxCeleb2 \cite{chung2018voxceleb2} as well as RGBD videos of 41 unseen identities from MMFace4D.


\textbf{Baselines} Our proposed method is the first neural approach attempting to create 3D facial animation driven by raw RGBD videos in an end-to-end manner, utilizing estimated optical flow fields to transform mesh vertices and deform the facial mesh. To provide a comprehensive evaluation of our method, we compare it with three state-of-the-art optical-flow-based 2D animation methods: FOMM \cite{siarohin2019first}, OSFV \cite{wang2021one} and DaGAN \cite{hong2022depth}. To animate RGBD images and drive the 3D mesh under our framework, we modify these methods and train them on the MMFace4D dataset using the loss function formulated in Sec. \ref{subsection:loss}. Both methods are initialized with pre-trained weights on VoxCeleb \cite{nagrani2017voxceleb}.


\begin{table}[t]
\centering
\caption{\textbf{Quantitative Results of Cross-identity Motion Retargeting.} We compared our method with three state-of-the-art methods using three designed tasks. The lower video FID \cite{wang2020g3an} indicates better generation qualities.}
\setlength{\tabcolsep}{8pt}
    \begin{tabular}{l|ccc}
    \toprule
    Method & Vox2$\rightarrow$Vox2 & MM$\rightarrow$Vox2 & MM$\rightarrow$MM \\
    \midrule 
    FOMM \cite{siarohin2019first} & 46.86        & 42.55      & 42.29    \\
    OSFV \cite{wang2021one} & 45.18        & 42.81      & 42.18    \\
    DaGAN \cite{hong2022depth} & 46.02 & 42.01 & 41.18 \\
    Ours & \textbf{44.47}        & \textbf{40.87}      & \textbf{40.69}  \\
    \bottomrule
    \end{tabular}
    \label{tab:trans}
\end{table}

\begin{figure}
    \centering
    \includegraphics[width=0.9\linewidth]{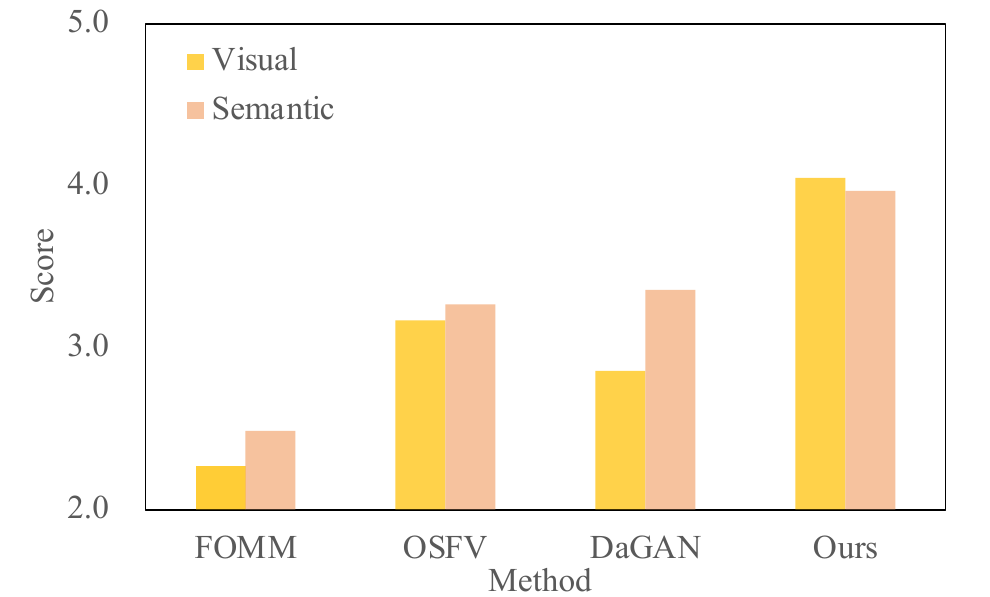}
    \caption{User Study of Motion Retargeting. We asked 20 users to evaluate the generated videos' visual quality and semantic consistency with the driving video. The score is in the range of 1-5, and a higher score denotes better. }
    \label{fig:user}
    \vspace{-10mm}
\end{figure}

\subsection{Evaluate Metrics}

We evaluate the performance of our model based on: (i) reconstruction fidelity using $\mathcal{L}_1$ and LPIPS metrics, (ii) generated video quality using video FID, (iii) semantic consistency using average keypoint distance (AKD), average Euclidean distance (AED) and cosine similarity (CSIM), and (iv) generated depth images quality using $\mathcal{L}_1$ and $\mathcal{L}_{sm}$ in Sec. \ref{subsection:loss}. These metrics provide us with a comprehensive evaluation of our model.



\textbf{Video FID} \cite{wang2020g3an}, derived from Fréchet inception distance (FID), is a metric that assesses both the visual quality and temporal consistency of the generated videos. Lower video FID indicates a higher quality of the generated videos. In our experiments, we utilize a pre-trained ResNext101 \cite{hara2018can} model to extract spatiotemporal features and compute video FID as an objective measure of video quality.

\textbf{AKD} aims to measure the difference between the facial landmarks of the reconstructed frame $\hat{\textbf{D}}$ and the real frame \textbf{D}. We adopt the facial landmark detection method proposed by Bulat and Tzimiropoulos \cite{bulat2017far} and compute the average distance between corresponding landmarks as AKD.

\textbf{AED and CSIM} \cite{zakharov2019few} both evaluate the ability to preserve identity while generating videos. We extracted identity embedded features with ArcFace \cite{deng2019arcface}, calculated the mean Euclidean distance between the identity embeddings as AED, and the cosine similarity between the embeddings as CSIM. 


\subsection{Quantitative Analysis}

To provide a quantitative analysis, we conduct two experiments to evaluate our framework thoroughly: same-identity reconstruction in Sec. \ref{subsubsection:recon} to assess the quality of our reconstruction, and cross-identity motion retargeting in Sec. \ref{subsubsection:trans} to evaluate the motion transfer ability of our approach.

\subsubsection{Same-identity Reconstruction}
\label{subsubsection:recon}

In this experiment, we aim to evaluate the reconstructing ability of our method. For simplicity, we focus on evaluating the quality of RGBD animation, as it directly affects the quality of mesh retargeting in our framework. We used the first frame as the source image (\textbf{S}) and the remaining frames as driving frames (\textbf{D}) to reconstruct the video. We conducted this experiment on the MMFace4D dataset and the VoxCeleb test set and reported the results in Tab. \ref{tab:recon}.

\begin{figure}
    \centering
    \includegraphics[width=1.0\linewidth]{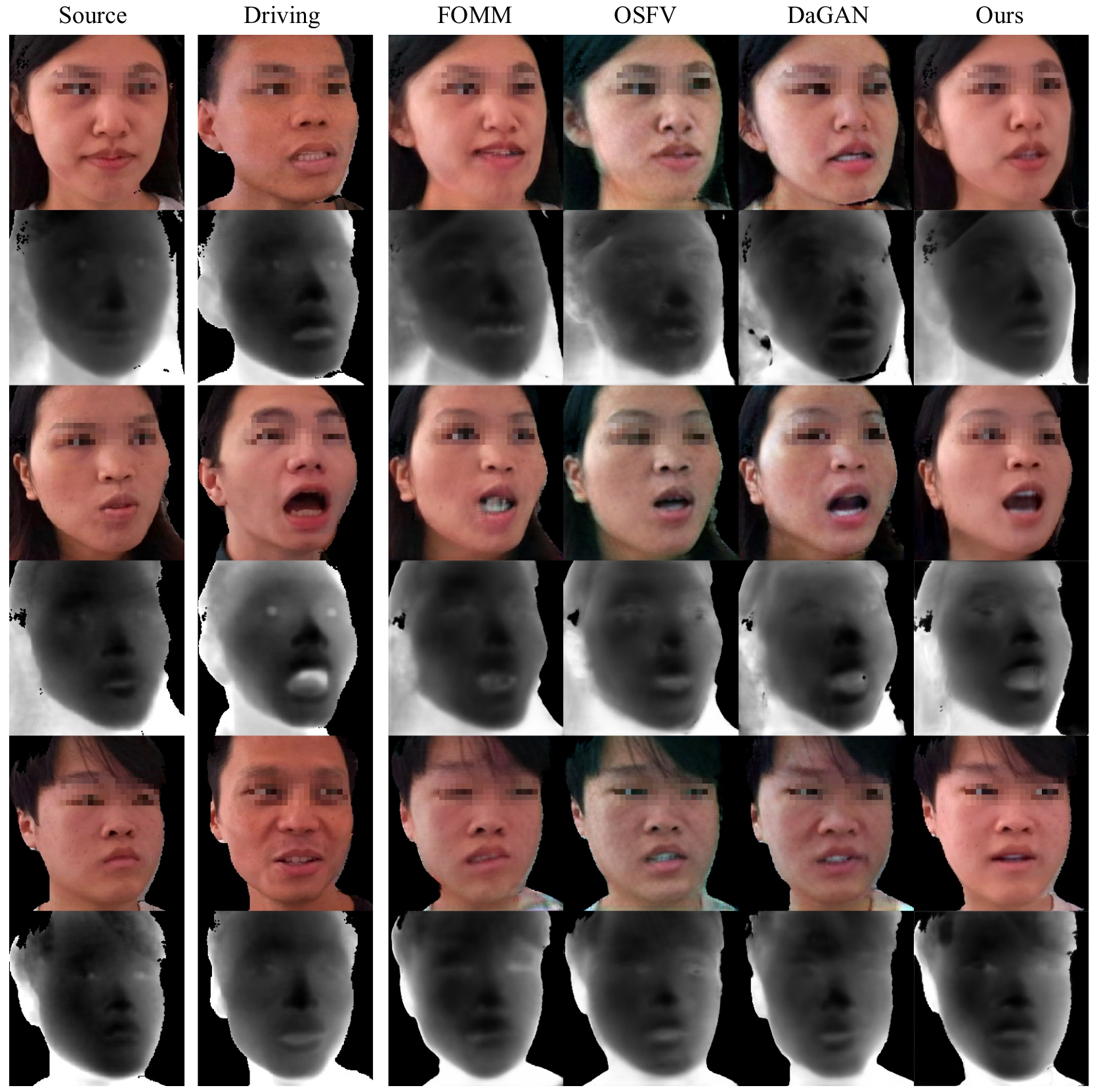}
    \caption{RGBD results of cross-identity motion retargeting. The first column shows the source images, while the second column shows the driving frames. The following columns show the transferred results of FOMM \cite{siarohin2019first}, OSFV \cite{wang2021one}, DaGAN \cite{hong2022depth}, and our method, respectively.  }
    \label{fig:transfer}
    \vspace{-6mm}
\end{figure}

As Tab. \ref{tab:recon} shows, our method achieves the best performance across all the metrics. Compared with the three baseline methods, our method achieves the highest reconstruction fidelity in both datasets, particularly in depth frames. This result further validates the effectiveness of the depth motion dictionaries proposed in this paper. While FOMM \cite{siarohin2019first} and OSFV \cite{wang2021one} treat the depth information as a simple image channel, and DaGAN \cite{hong2022depth} fails to model motion in the depth field, our method excels in depth reconstruction due to the depth motion dictionaries. Furthermore, our method achieves the highest scores in AKD, AED, and CSIM, indicating its ability to transfer motion while preserving the identity of the source character. These results highlight the strength of our multi-level flow-based generator.

\begin{figure}[t]
\centering
    \includegraphics[width=0.8\linewidth]{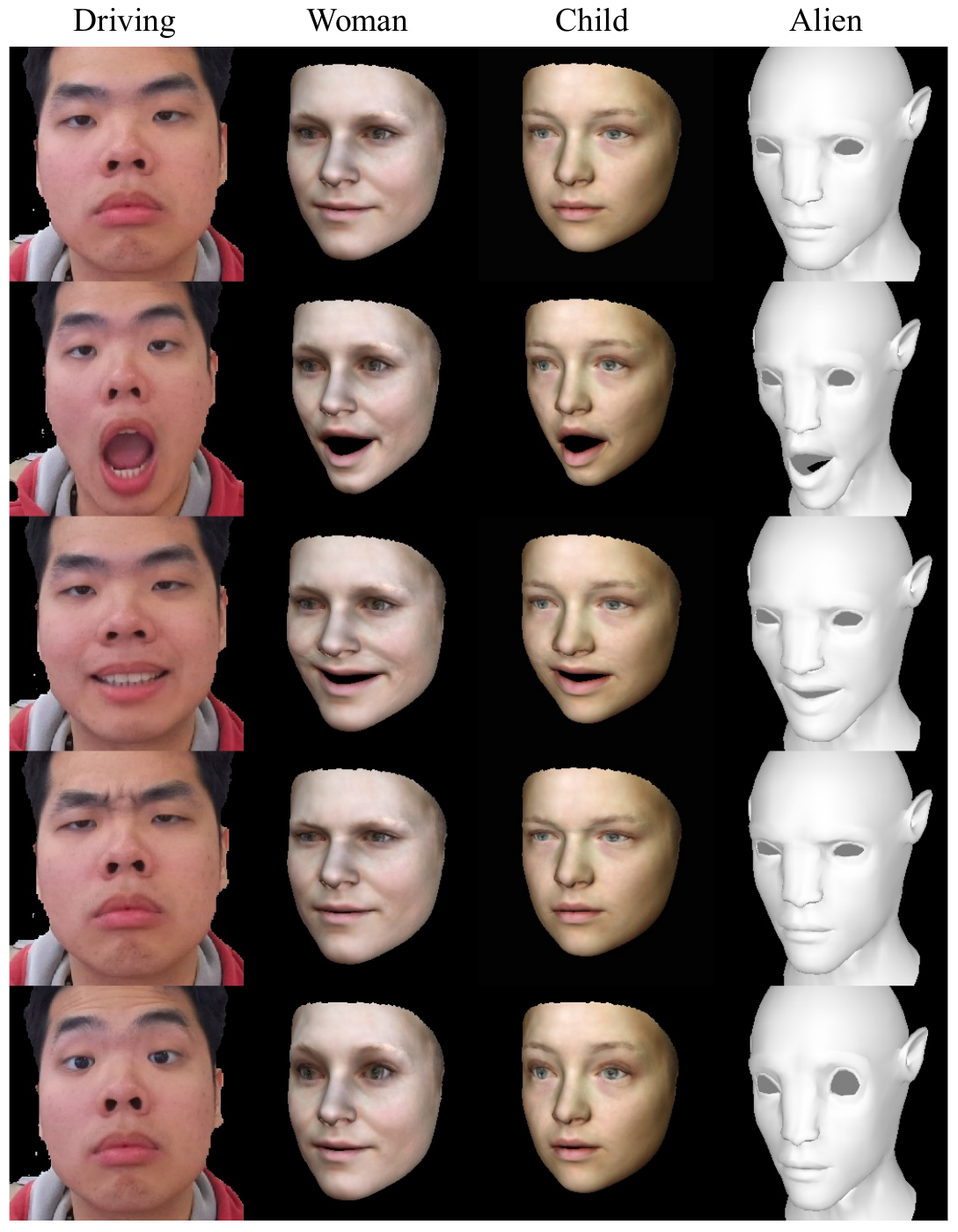}
    \caption{Qualitative results from our method. The leftmost column displays the driving frames. In contrast, the subsequent columns exhibit three target characters: a woman, a child, and an alien. The top row shows the source meshes. More results are presented in the supplementary material. }
    \label{fig:results}
    \vspace{-6mm}
\end{figure}

\subsubsection{Cross-identity Motion Retargeting }
\label{subsubsection:trans}

In this experiment, we aim to assess the motion transfer ability of our method. Specifically, we used source images (\textbf{S}) and driving frames (\textbf{D}) from different video sequences, which differs from Sec. \ref{subsubsection:recon}. We designed three tasks: driving source images from VoxCeleb2 with videos from VoxCeleb2 (Vox2$\rightarrow$Vox2), driving source images from MMFace4D with videos from VoxCeleb2 (MM$\rightarrow$Vox2), and driving source images from MMFace4D with videos from MMFace4D (MM$\rightarrow$MM). It's important to note that all the images and videos used here were unseen during the training of our models, ensuring a fair evaluation.

As the ground truth animation videos were unavailable, we used video FID \cite{zhang2018unreasonable} to assess our generated videos' visual quality and temporal consistency. We randomly selected 2200 source images and driving video clips for each task to generate retargeted videos. These videos were then downsampled to the resolution of $112\times 112$ and randomly cut to 32 frames. We computed video FID by calculating the distance between the generated data and the real data distributions sampled from the source dataset. 
The results are presented in Tab. \ref{tab:trans}. Our method consistently outperforms the other methods regarding video FID for all tested tasks, demonstrating superior motion transfer ability. 
To provide an intuitive demonstration of the performance of the four methods, we show some transferred RGBD results in Fig.~\ref{fig:transfer}. FOMM produced some artifacts, such as a puffy face, while OSFV generated noisy results in color and depth frames. Although DaGAN transferred the facial motion better, it did not preserve the identity well. In contrast, our method generated the most natural and clearest color frame and the cleanest and smoothest depth frame, achieving the best performance in transferring the facial motion of RGBD frames. 

To further compare the effectiveness of our method with baseline methods, we conducted a user study. Each participant was asked to evaluate and rate the videos generated by the methods. Specifically, we randomly generated groups of videos. Each video group contained a video generated by our method and three videos generated by the three baselines. These video groups and their corresponding driving videos were presented to 20 human raters. The raters were asked to evaluate the video's visual quality and semantic consistency. 
As Fig. \ref{fig:user} reported, our method obtained the highest scores, which means that our method generates the most realistic video while transferring facial motion from driving videos. 

Notably, the three baselines rely on facial keypoints transformation, so their performance may be affected by the accuracy of the keypoint detector. However, our method captures facial motion by hierarchical motion dictionaries and generates RGBD frames coarse-to-fine, which facilitates realistic motion retargeting.

\subsection{Qualitative Analysis}
\label{subsection:res}

In Figure \ref{fig:results}, we present qualitative examples of our proposed method. Specifically, we recorded an RGBD video using the Azure Kinect V2 to drive the facial expressions of three target characters: a woman, a child, and an alien. Despite the dissimilarity between the actor and the target characters, our method generated impressive results and accurately retargeted facial motion, particularly the motion of the mouth, and transferred micro-expressions, such as eye-widening, squinting, and mouth stretching. Furthermore, our method demonstrated impressive ability in animating the alien avatar, which had a significantly different appearance from the actor and was not seen during the training phase. However, expressions such as rolling eyes, gazing, and sticking out the tongue could not be transferred to the target character, as the target mesh did not model eyes and tongue separately. Overall, our results demonstrate the potential of our method as a novel solution for generating 3D facial animation.

\begin{figure}[t]
\centering
    \subfigure[Wave-lip artifact]{
        \begin{minipage}[t]{1.3in}
        \centering
        \includegraphics[width=1.2in]{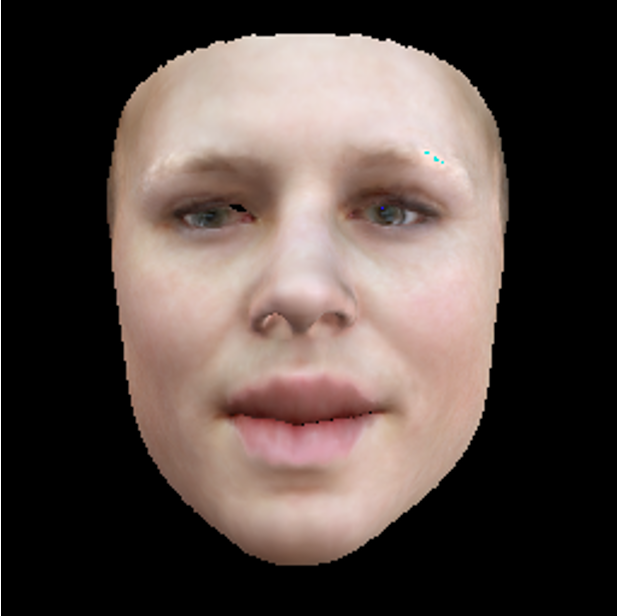}
        \end{minipage}
        \label{fig:wave-lip}
    }
    \subfigure[Our method]{
        \begin{minipage}[t]{1.3in}
        \centering
        \includegraphics[width=1.2in]{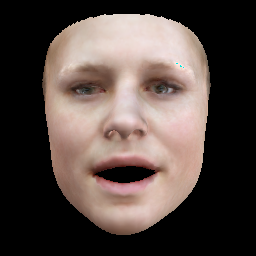}
        \end{minipage}
        \label{fig:geod}
    }
    \caption{Comparison of mesh retargeting results. (a) Wave-lip artifacts are caused by using Euclidean distance to calculate blend weights $w_{i,j}$. (b) More natural results are obtained with our method using geodesic distance to determine $w_{i,j}$.}
    \label{fig:warp}
    \vspace{-6mm}
\end{figure}

\subsection{Ablation Study}
\label{subsection:ablation}

\subsubsection{Controlling Weights Calculation}
As discussed in Section \ref{subsection:retarget}, using geodesic distances to calculate controlling weights is crucial for producing accurate retargeted results. To further verify this assertion, we present a case study. When controlling weights are calculated using Euclidean distance, artifacts such as the wave-lip artifact can occur when the mouth is open, as illustrated in Fig. \ref{fig:wave-lip}. 
This is because the movement of controlling points from the lower lip heavily influences the vertex of the upper lip. However, calculating blend weights using geodesic distance can avoid such artifacts, as the controlling points of the lower lip will not be considered neighbors of the vertex in the upper lip. Therefore, our method generates more natural results, as shown in Fig. \ref{fig:geod}.

\begin{table}[t]
\centering
\setlength{\tabcolsep}{6pt}
\caption{Ablation Study on Depth Motion Dictionary.}
    \begin{tabular}{l|ccc|cc}
    \toprule
    \multicolumn{1}{c|}{\multirow{2}{*}{Size of $D_i^{depth}$}} & \multicolumn{3}{c|}{MMFace4D} & \multicolumn{2}{c}{VoxCeleb} \\
    \multicolumn{1}{c|}{}   
    & $\mathcal{L}_1$     & LPIPS   & Depth $\mathcal{L}_1$  & $\mathcal{L}_1$            & LPIPS         \\
    \cmidrule(lr){1-6}
    0                                           & 8.64   & 0.118   & 0.043     & 11.42         & 0.129         \\
    5                                           & \textbf{8.04}   & \textbf{0.104}   & \textbf{0.030}     & \textbf{11.26}         & \textbf{0.119}         \\
    10                                          & 8.71   & 0.116   & 0.036     & 11.69         & 0.126   \\
    20                                          & 8.73   & 0.119   & 0.035     &   11.35       &  0.127  \\
    \bottomrule
    \end{tabular}
    \label{tab:abl}
    \vspace{-4mm}
\end{table}

\subsubsection{Depth motion dictionary}
We provide an in-depth analysis of our design of the depth motion dictionaries $\{D_i^{depth}\}_1^6$ in the generator, as discussed in Sec. \ref{subsubsection:gen}. We focus on whether introducing $D_i^{depth}$ benefits the generation of RGBD frames and determine the optimal number of basis vectors that $D_i^{depth}$ requires. 

Here we performed the same task as discussed in Sec. \ref{subsubsection:recon}, and reported the reconstruction faithfulness metrics, \textit{i.e.}, $\mathcal{L}_1$, and LPIPS. As shown in Tab. \ref{tab:abl}, the depth motion dictionary $D_i^{depth}$ indeed benefits the reconstruction ability of our method, especially in terms of depth image generation. Notably, when the size of $D_i^{depth}$ is set to  5, our model achieves the best reconstruction results, which indicates that a few basic transformations can represent the depth motion space. Thus, a small depth motion dictionary is sufficient to model facial motion in the depth field.

\section{Conclusion}
In this paper, we propose a novel self-supervised framework, Versatile Face Animator, for transferring facial motion from captured RGBD videos to 3D facial meshes to create 3D facial animation. Our framework comprises two modules: a flow-based RGBD animation module that animates RGBD frames with hierarchical motion dictionaries and a mesh retarget module that performs 3D facial retargeting using blend transformations. Our end-to-end approach eliminates the need for labor-intensive and time-consuming blendshape-based methods or facial rigging techniques. Extensive experiments demonstrate that our framework is a promising and cost-efficient solution for generating 3D facial animation compared with existing literature. However, there are still some limitations to our method. The RGBD animation module may not perform well in some occluded cases, and more training data may be required to improve retarget performance for unseen avatars. Additionally, the estimation of the controller transformations and the accuracy of the generated depth frames significantly influence the realisticness of the retargeted mesh. In future work, we plan to focus on improving the quality of generated RGBD frames and the versatility of our framework for 3D facial animation production. We believe that the simplicity, efficiency, and versatility of our framework are crucial steps toward the future of the metaverse.

\section{Acknowledgements}
This work is supported by the National Key R\&D Program of China under Grant No. 2021QY1500, the State Key Program of the National Natural Science Foundation of China (NSFC) (No.61831022).

\newpage
\bibliographystyle{ACM-Reference-Format}
\balance
\bibliography{acmart}

\end{document}